%% file: main.tex
\documentclass[sigconf,nonacm,numbers]{acmart}

\usepackage{csquotes}

\usepackage[toc=false,acronym,symbols,shortcuts=ac]{glossaries-extra}
\setabbreviationstyle[acronym]{long-short}
\glsdisablehyper
\loadglsentries{glossary}

\usepackage{paralist}

\usepackage[subrefformat=parens,labelformat=parens]{subcaption}
\usepackage{makecell}
\newcommand{\tabitem}{~~\llap{\textbullet}~~}

\usepackage[capitalise,nameinlink]{cleveref}

\begin{document}

\title{Informed, but Not Always Improved: Challenging the Benefit of Background Knowledge in GNNs}
\titlenote{This work was presented at the 24th International Workshop on Data Mining in Bioinformatics (BIOKDD 2025), held on August 3–7, 2025, in Toronto, Canada.}

\author{Kutalmış Coşkun}
\affiliation{%
  \institution{University of Rostock}
  \city{Rostock}
  \country{Germany}
}
\email{kutalmis.coskun@uni-rostock.de}

\author{Ivo Kavisanczki}
\affiliation{%
  \institution{University of Rostock}
  \city{Rostock}
  \country{Germany}
}
\email{ivo.kavisanczki@uni-rostock.de}

\author{Amin Mirzaei}
\affiliation{%
  \institution{University of Rostock}
  \city{Rostock}
  \country{Germany}
}
\email{mohammad.mirzaei@uni-rostock.de}

\author{Tom Siegl}
\affiliation{%
  \institution{University of Rostock}
  \city{Rostock}
  \country{Germany}
}
\email{tom.siegl@uni-rostock.de}

\author{Bjarne C. Hiller}
\affiliation{%
  \institution{University of Rostock}
  \city{Rostock}
  \country{Germany}
}
\email{bjarne.hiller@uni-rostock.de}

\author{Stefan Lüdtke}
\affiliation{%
  \institution{University of Rostock}
  \city{Rostock}
  \country{Germany}
}
\email{stefan.luedtke@uni-rostock.de}

\author{Martin Becker}
\affiliation{%
  \institution{University of Rostock, University of Marburg}
  \city{Marburg}
  \country{Germany}
}
\email{martin.becker@uni-marburg.de}

\renewcommand{\shortauthors}{Coşkun et al.}

\begin{abstract}
    In complex and low-data domains such as biomedical research, incorporating background knowledge (BK) graphs, such as protein-protein interaction (PPI) networks, into graph-based machine learning pipelines is a promising research direction.
    However, while BK is often assumed to improve model performance, its actual contribution and the impact of imperfect knowledge remain poorly understood.
    In this work, we investigate the role of BK in an important real-world task: cancer subtype classification.
    Surprisingly, we find that (i) state-of-the-art GNNs using BK perform no better than uninformed models like linear regression, and (ii) their performance remains largely unchanged even when the BK graph is heavily perturbed.
    To understand these unexpected results, we introduce an evaluation framework, which employs (i) a synthetic setting where the BK is clearly informative and (ii) a set of perturbations that simulate various imperfections in BK graphs. 
    With this, we test the robustness of BK-aware models in both synthetic and real-world biomedical settings.
    Our findings reveal that careful alignment of GNN architectures and BK characteristics is necessary but holds the potential for significant performance improvements.
\end{abstract}

\maketitle

\section{Introduction}

\Ac{BK}, or prior knowledge, refers to domain-specific information that can enhance \ac{ML} models.
Incorporating BK --- a core concept in \ac{IML} \cite{vonruedenInformedMachineLearning2023} --- can improve performance by providing structural context beyond what is learned from data alone.
This is especially useful in low-data settings, where BK constrains the hypothesis space and guides learning toward physically/biologically plausible solutions, or in applications demanding explainability where BK offers interpretable, domain-relevant reasoning paths \cite{sirocchiMedicalinformedMachineLearning2024, vonruedenInformedMachineLearning2023}.
\ac{IML} has been applied in diverse fields, including physics-guided learning \cite{dawPhysicsguidedNeuralNetworks2022}, immunology \cite{culosIntegrationMechanisticImmunological2020}, drug discovery \cite{jiangPocketFlowDataandknowledgedrivenStructurebased2024}, and medical imaging \cite{wangLifespangeneralizableSkullstrippingModel2025}.
A common way to represent BK is through structured formats such as graphs.
In biology and biomedicine, common graph-based BK sources include knowledge graphs \cite{chandakBuildingKnowledgeGraph2023}, \ac{GGI}/\ac{PPI} networks \cite{szklarczykSTRINGV11Protein2019, kamburovConsensusPathDBMoreComplete2011,oughtredBioGRIDDatabaseComprehensive2021}, and pathway databases \cite{milacicReactomePathwayKnowledgebase2024}.
These graphs capture functional, regulatory, or physical dependencies that are not directly apparent in expression data, thus generally assumed to provide valuable context that supports various predictive tasks.

\Acp{GNN} provide a natural way to leverage such structured data, enabling the direct incorporation of BK into \ac{ML} pipelines.
In biomedical research, GNNs have been particularly effective in
applications such as drug-drug interaction prediction \cite{linKGNNKnowledgeGraph2020}, single-cell classification \cite{wangAdversarialDenseGraph2023} and cancer gene prediction \cite{schulte-sasseIntegrationMultiomicsData2021}.
These successes highlight the potential of BK-enhanced GNNs in complex real-world tasks.

Despite promising results, it remains unclear when and how BK meaningfully benefits GNN-based models.
The effectiveness of BK likely depends on factors such as graph structure, model architecture, and data characteristics.
Moreover, the robustness of GNNs to imperfections in BK graphs has not been systematically examined.
A deeper understanding of these aspects is essential for reliably applying BK-enhanced GNNs in real-world settings.

In this work, we investigate the role of BK in an important real-world task: cancer subtype classification using \ac{GGI}/\ac{PPI} networks.
This setting is representative of how BK is commonly used in biomedical research.
However, we uncover two surprising results: 
\begin{inparaenum}[(i)]
    \item simple, uninformed baselines such as \ac{LogReg} outperform state-of-the-art GNNs, and
    \item their performance remains largely unchanged even when the BK graph is heavily perturbed or entirely removed.
\end{inparaenum}
This unexpected insensitivity raises the question of whether GNNs are truly leveraging BK in practice.

To better understand this result, we design a controlled synthetic setting where BK is constructed to be clearly informative.
This setting serves as an initial verification that GNNs can, in principle, leverage useful graph structure, and to probe the conditions under which BK contributes meaningfully to model performance.
In addition, we define a set of perturbation methods to systematically examine how GNNs respond to various types of imperfections in BK graphs.

Overall, our key contributions are as follows:
\begin{itemize}
    \item We demonstrate that existing methods do not effectively utilize BK graphs.
    This challenges common assumptions and motivates the need for improved integration techniques and systematic analysis to measure BK usefulness.
    \item We create a synthetic setting which shows that GNNs can theoretically benefit significantly from structured BK graphs.
    These datasets serve as controlled benchmarks for testing when and how BK can, in principle, provide an advantage.
    \item We show in the synthetic setting that GNNs are more tolerant of small amounts of random, uniformly distributed errors than systematic errors in BK, revealing the risk of relying on biased BK sources.
    \item We find initial indicators that, GNNs are more robust to missing information than to incorrect information in the BK, emphasizing BK correctness may be more important than completeness.
\end{itemize}

\section{Related Work}
\label{sec:rw}

Integration of \ac{BK} into \ac{ML} models is a growing area of research.
While much work has explored how to incorporate \ac{BK}, particularly in \acp{GNN}, there has been comparatively little attention to how robust these models are when the \ac{BK} is faulty, incomplete, or biased.
This section surveys related research across three areas:
\begin{inparaenum}
    \item \ac{IML} with imperfect \ac{BK}
    \item the integration of \ac{BK} in \acp{GNN}, and
    \item adversarial attacks targeting \acp{GNN}.
\end{inparaenum}

\subsection{IML with Imperfect BK}

A central challenge in leveraging \ac{BK} for \ac{ML} models lies in ensuring the quality and reliability of the provided information.
While much of the existing work focuses on incorporating \ac{BK} into learning pipelines, a smaller body of research has examined how robust these models are when the \ac{BK} is incomplete, noisy, or erroneous.
For instance, the \ac{iEN} model \cite{culosIntegrationMechanisticImmunological2020} incorporates expert knowledge as feature-specific regularization weights within a regularized ElasticNet framework.
The authors further explored the model’s resilience by incrementally introducing noise to these regularization weights and assessing whether performance still benefited from the \ac{BK}.
Another example, SimKern \cite{deistSimulationassistedMachineLearning2019}, integrates knowledge in the form of \acp{ODE} used for simulations.
In addition to comparing against uninformed baselines, this study also evaluated the model's performance when using two distinct sets of \acp{ODE} with varying quality.

In general, many studies in this area rely, if at all, on simple ablation studies that either remove the \ac{BK} or compare the proposed model to an uninformed baseline (e.g., \cite{dawPhysicsguidedNeuralNetworks2022, wangLifespangeneralizableSkullstrippingModel2025}).
However, such evaluations are typically limited in scope and do not involve \acp{GNN}.
As a result, a systematic investigation into the robustness of \acp{GNN} to faulty or imperfect \ac{BK} remains largely unexplored.

\subsection{Integration of BK into GNNs}

GNNs that incorporate BK have been applied to a range of tasks in bioinformatics, including cancer subtype classification.
Notable examples include the multimodal GNN proposed by \citet{liMultimodalGraphNeural2024} and MPK-GNN \cite{xiaoGraphNeuralNetworks2023}, both of which are evaluated in this study and will be introduced in detail in \cref{sec:bg_csc}.
However, neither model includes a systematic analysis of robustness to perturbations in the BK graphs.
An extension of MPK-GNN, MKI-GNN \cite{huangMultilevelKnowledgeIntegration2023}, integrates multiple BK graphs, yet it similarly lacks an evaluation of how effectively the model leverages each graph or whether it is resilient to flawed or noisy knowledge.

Another relevant application in bioinformatics is cancer gene detection, which aims to identify genes that play a critical role in cancer development.
In this setting, BK is often provided in the form of \ac{PPI} networks.
The EMOGI model \cite{schulte-sasseIntegrationMultiomicsData2021}, for example, employs PPI data from ConsensusPathDB \cite{kamburovConsensusPathDBMoreComplete2011} for semi-supervised node classification.
To assess the role of BK, the authors tested the model with different \ac{PPI} graphs and applied targeted perturbations.
One such perturbation involved a double edge swap, where two randomly selected edges $(x,y)$ and $(u,v)$ are replaced with $(x,u)$ and $(y,v)$, preserving the degree distribution of the nodes.
This process was applied incrementally until all edges were swapped.
Additionally, they evaluated performance using a synthetic graph generated via the Holme–Kim algorithm \cite{holmeGrowingScalefreeNetworks2002}, which produces graphs with power-law degree distributions, a characteristic of biological networks \cite{barabasiEmergenceScalingRandom1999}.
These experiments showed a substantial decline in model performance under graph perturbation, suggesting that the model relies on the \ac{PPI} information.
However, these findings are limited to a single GNN architecture, and systematic studies across multiple models and types of BK degradation remain scarce.

Finally, \citet{bertinAnalysisGeneInteraction2020} examined the adequacy of \ac{GGI} networks as BK in \ac{ML} methods.
The framework introduced in \cite{bertinAnalysisGeneInteraction2020} defines a GGI network as useful if the expression level of a target gene can be accurately predicted from the neighbor genes on the graph, compared to using all genes as predictors.
This analysis showed that STRINGDB satisfies this criterion.
However, the framework only considers prediction of expression levels, and does not evaluate the utility of BK for more complex tasks.

\subsection{Adversarial Attacks Targeting GNNs}

A related research area is adversarial attacks on GNNs, which involve manipulating input data to induce misclassifications.
One prominent example is Nettack \cite{zugnerAdversarialAttacksGraph2020}, an algorithm that strategically alters graph structures and node features to mislead GNNs in node classification tasks.
It does so by identifying effective edge insertions, deletions, or feature modifications using a surrogate model.
Other methods include NIPA \cite{sunAdversarialAttacksGraph2020}, which injects fake nodes and links, and the gradient-based attack by \citet{xuTopologyAttackDefense2019}, which perturbs the graph by modifying its topology.

However, these approaches target the integrity of input data rather than the quality of BK.
As such, they do not address the robustness of GNNs that explicitly incorporate BK when that knowledge is faulty or biased.

\section{Background}
\label{sec:bg}

This section provides an overview of relevant concepts and prior work related to \ac{IML} and \acp{GNN}.

\subsection{Informed Machine Learning (IML)}

\Ac{IML} incorporates \ac{BK} to guide models beyond purely data-driven learning \cite{vonruedenInformedMachineLearning2023}.
There is a variety of \ac{BK} modalities, and there are several ways to incorporate \ac{BK} into \ac{ML} models:
\ac{BK} can shape the training data (e.g., via data augmentation), inform the model architecture (e.g., by introducing specialized layers), guide the training process (e.g., through regularization), or be used to validate model outputs under specific constraints or conditions \cite{vonruedenInformedMachineLearning2023}.

\subsection{Graph Neural Networks (GNNs)}
\label{sec:gnns}

\Acp{GNN} are a class of neural architectures tailored for learning from data represented as graphs.
Graphs naturally model relational data and are widely used in biomedical research.
Unlike traditional neural networks that operate on fixed-structure data such as images or tabular inputs, \acp{GNN} exploit the relational structure inherent in graphs to learn more expressive representations.
The core mechanism of a GNN is message passing \cite{gilmerNeuralMessagePassing2017}, where each node iteratively aggregates information from its neighbors and combines it with its own features to update its representation.
These \enquote{messages} are typically the feature vectors of neighboring nodes, passed along graph edges.
This process is performed iteratively over multiple layers, each layer enabling a node to incorporate information from increasingly distant neighbors.
With $k$ layers, a GNN enables each node to incorporate information from nodes up to $k$ hops away.

\subsection{Cancer Subtype Classification Models}
\label{sec:bg_csc}

In this section, we provide an overview of the two informed \ac{GNN} models evaluated in this study: 
\begin{inparaenum}
    \item the model developed by \citet{liMultimodalGraphNeural2024} (referred to as Li24), and
    \item \ac{MPK-GNN} \cite{xiaoGraphNeuralNetworks2023}.
\end{inparaenum}
Both models address cancer subtype classification, a crucial task in precision oncology that enables personalized treatment and provides deeper insights into tumor heterogeneity.
These two recent models (2023–2024) are selected as benchmarks due to their strong reported results on this task.
The performance of these models on original and perturbed BK graphs is analyzed in \cref{sec:results_real}, where we assess their robustness to changes in network structure and compare them to the uninformed baselines.

\subsubsection{Li24}
\label{sec:bg_li24}

Li24 \cite{liMultimodalGraphNeural2024} was specifically developed for molecular cancer subtype classification.
It performs graph classification by integrating multi-omics data with a biological \ac{BK} graph.
The model incorporates three types of -omics data: gene expression (mRNA), copy number variation (CNV), and microRNA (miRNA) profiles.
Gene expression data quantify how actively each gene is expressed across samples, while CNV data reflect gene dosage by indicating the number of copies present.
miRNA data capture the expression patterns of individual microRNAs.
Each data type is structured as a matrix, with samples as rows and genes or microRNAs as columns.

\paragraph{Biological Graph}
The model operates on a composite biological graph, or \emph{supra-graph}, shared across all samples, with node features varying by sample.
The graph contains two types of nodes: genes and miRNAs, while edges are derived from two biological data sources: gene-gene edges are obtained from the BioGRID gene interaction network \cite{oughtredBioGRIDDatabaseComprehensive2021}, and miRNA-gene edges are sourced from the miRDB target network \cite{chenMiRDBOnlineDatabase2020}.
In addition, inferred miRNA–miRNA edges are added between miRNAs that share at least one common target gene.
All edges are undirected and weighted, with weights reflecting the reliability of interactions, quantified by the number of supporting scientific publications.

\paragraph{Model Architecture}
The Li24 model employs a parallel architecture that combines a \ac{GNN} with a \ac{MLP} to capture both local and global patterns in the data. 
The \ac{MLP} branch processes the multi-omics data independently to extract global signals, while the \ac{GNN} branch models local interactions within the biological graph (i.e., BK and the multi-omics data).
For the \ac{GNN}, both \ac{GCN} and \ac{GAT} layers are evaluated.
Finally, representations from both branches are concatenated and fed into a classification layer.

\paragraph{Evaluation}
Li24 was tested on two datasets from \ac{TCGA}: \ac{PANCAN} and \ac{BRCA}. 
The results show that Li24 outperforms both a standalone \ac{MLP} and other state-of-the-art \ac{GNN}-based models.
On the \ac{BRCA} dataset with 700 genes, the \ac{GAT}-based variant achieved an accuracy of $86.4\% \pm 1.9\%$, exceeding the \ac{MLP} baseline by $5.6\%$.
On \ac{PANCAN} with 700 genes, it reached $83.9\% \pm 1.4\%$, with a $5.5\%$ improvement over the \ac{MLP}.

\subsubsection{MPK-GNN}

\ac{MPK-GNN} \cite{xiaoGraphNeuralNetworks2023} introduces a contrastive learning framework for integrating multiple sources of \ac{BK} in the form of graphs.
The model is initially designed to be task-agnostic, learning a general-purpose embedding that can later be fine-tuned or incorporated into task-specific modules.
For evaluation, \ac{MPK-GNN} was applied to molecular cancer subtype classification, aligning with the experimental setup of Li24 \cite{liMultimodalGraphNeural2024} described in \cref{sec:bg_li24}.

\paragraph{Multiple BK Graphs}
In contrast to Li24, which builds a single supra-graph from multiple data sources, \ac{MPK-GNN} learns directly from multiple \ac{BK} graphs, each capturing distinct biological relationships between genes or proteins.
In all graphs, nodes represent genes, while edges (undirected and weighted) are drawn from diverse sources: \ac{GGI} from BioGRID \cite{oughtredBioGRIDDatabaseComprehensive2021}, \ac{PPI} and co-expression data from STRINGDB \cite{szklarczykSTRINGV11Protein2019}.
Node features are derived from multi-omics data, which serve as input for downstream learning.

\paragraph{Model Architecture}
The architecture of \ac{MPK-GNN} is composed of several modules, each targeting different facets of the input data.
A sample-level module, implemented as an \ac{MLP}, captures global patterns from the multi-omics features.
In parallel, a feature-level module employs a \ac{GCN} with shared weights to process multiple \ac{BK} graphs.
Each graph is passed through the \ac{GCN} to generate an individual embedding, which is then refined by a projection module that prepares the representations for contrastive learning.
A task-specific module then integrates the sample-level embedding with the graph embeddings to perform the final classification.
In addition to a task-specific loss (such as cross-entropy for classification), \ac{MPK-GNN} also employs a contrastive loss, which encourages consistency across the embeddings from different \ac{BK} graphs. 

\paragraph{Evaluation}
\ac{MPK-GNN} achieved $84.0\% \pm 0.4\%$ accuracy on \ac{PANCAN}, surpassing state-of-the-art CMSC with $82.9\% \pm 0.4\%$.

\section{The Test Framework}
\label{sec:method}

This section introduces how we assess the BK utilization in GNNs\footnote{\url{https://bckrlab.org/p/kill-gnn}}.

\subsection{BK Perturbation Methods}
\label{sec:perturbations}

To investigate how \acp{GNN} respond to imperfect \ac{BK}, we introduce controlled perturbations to the \ac{BK} graphs.
These perturbation types simulate two primary types of knowledge imperfections: missing and incorrect information.
Missing knowledge corresponds to relevant nodes or edges being absent from the graph, while incorrect knowledge introduces false or misleading connections.
In addition to type, perturbations differ in their distribution.
Random perturbations affect the graph in an unstructured manner to capture unpredictable noise or uncertainty often present in real-world data.
In contrast, systematic perturbations target specific nodes, edges, or substructures, modeling structured biases.
For example, systematic omissions may arise from under-studied biological processes, whereas systematic inaccuracies can reflect flaws in experimental design or curation.
The interpretation and impact of these perturbations depend on the structure and purpose of the \ac{BK} graph.
An overview of the types of applied perturbations is given in \cref{tab:perturbations}, and they are explained in more detail in \cref{sec:pert_rem_add,sec:pert_noise,sec:pert_det,sec:pert_move}.
\begin{table}[htbp]
    \caption{Overview of applied BK graph perturbations.}
    \label{tab:perturbations}
    \begin{tabular}{lll}
        \toprule
        & \textbf{Missing} & \textbf{Incorrect} \\ \midrule
        \textbf{Random} & \tabitem Removing edges & \tabitem Adding edges \\ \midrule
        \textbf{Systematic} & \makecell[cl]{\tabitem Isolating nodes \\\tabitem Edge weight noise\\(negatives removed)} & \makecell[cl]{\tabitem Detach-and-Rewire \\\tabitem Edge weight noise\\(negatives replaced)} \\
        \bottomrule
    \end{tabular}
\end{table}

To measure the severity of perturbations, we introduce the parameter $\kappa$, where $\kappa=0$ corresponds to an unaltered graph.
The meaning of $\kappa$ depends on the type of perturbation and is explained in the corresponding section.

\subsubsection{Removing and Adding Edges}
\label{sec:pert_rem_add}

A straightforward method for introducing perturbation is to either randomly remove or add edges in the \ac{BK} graph, corresponding to uniformly distributed missing and incorrect knowledge, respectively.
In this case, the severity measures $\kappa_{-}$ and $\kappa_{+}$ correspond to the ratios given in \cref{eq:kappa_remove_add}.
\begin{equation}
    \label{eq:kappa_remove_add}
    \kappa_{-}=\frac{\text{\#removed edges}}{|E|}, \qquad \kappa_{+}=\frac{\text{\#added edges}}{|E|}
\end{equation}

For edge removal, $\kappa_{-}$ takes values from $0$ (no removal) and $1$ (all edges are removed).
Similarly, for adding edges $\kappa_{+}=0$ indicates no edges are added and while values greater than $1$ is possible, $\kappa_{+}=1$ means the number of edges is doubled.
Sufficiently high values of $\kappa_{+}$ can yield a fully connected network.

\subsubsection{Edge Weight Noise}
\label{sec:pert_noise}

The next form of perturbation introduces noise to the edge weights of the BK graph and captures aspects of both missing and incorrect information.
It is applied to the \ac{BK} graph from BioGrid, which is used in \cite{liMultimodalGraphNeural2024}.
In BioGrid, edge weights represent evidence scores, quantifying the number of scientific publications that report a given interaction \cite{oughtredBioGRIDDatabaseComprehensive2021}.

The core idea behind this method is that edges with lower evidence scores are more likely to be missing or unreliable than those with higher scores.
By perturbing edge weights in proportion to their evidence level, this approach aims to simulate error patterns that are more representative of real-world uncertainties.

Noise is drawn from a zero-mean normal distribution with standard deviation $\kappa_{\sigma}$, and is rounded to the nearest integer to match the integer-valued edge weights.
If the added noise results in a negative edge weight, the two variants of this perturbation method either remove or randomly replace the corresponding edge.
This outcome is more likely for edges with initially low weights, modeling the greater uncertainty associated with low-evidence interactions.
Although the noise is sampled randomly, the perturbation is systematic in nature, as it disproportionately affects low-confidence edges.
In practice, this perturbation method 
\begin{inparaenum}[(i)]
    \item adds noise to edges, and
    \item probabilistically removes/replaces low evidence edges.
\end{inparaenum}

The variant that replaces edges with negative weights simulates incorrect knowledge that may exist in the BK graph (i.e., spurious or non-existent interactions).
The weight of the new edge is sampled from the empirical distribution of existing edge weights.
This approach enables the modeling of incorrect information with a single perturbation parameter, $\kappa_{\sigma}$.

\subsubsection{Isolating Nodes}
\label{sec:pert_det}

Isolating a node removes all edges attached to it.
This operation simulates missing knowledge, as the node itself remains present, but its interactions are entirely unobserved.
Depending on the selection procedure of nodes to isolate, it can simulate random or systematic errors. 
The severity of this perturbation $\kappa_{i}$ is measured as the ratio of nodes isolated.

\subsubsection{Detach-and-Rewire}
\label{sec:pert_move}

This perturbation method is designed for scenarios where the \ac{BK} graph contains clusters.
It simulates incorrect knowledge by isolating a subset of nodes from each cluster (via edge removal) and rewiring them to a different cluster.
This operation misassigns nodes to incorrect functional groups, modeling structured but erroneous associations while controlling the extent via the parameter $\kappa_{dr}$, which is ratio of nodes affected.
While applicable to any \ac{BK} graph with identifiable clusters, this perturbation is specifically designed for the synthetic setting introduced in \cref{sec:synthetic}, where the connections are known to be informative.

\subsection{Synthetic Setting: Ensuring BK Informativeness}
\label{sec:synthetic}

To evaluate whether \acp{GNN} can effectively leverage information from a \ac{BK} graph, we construct a controlled synthetic scenario that caters to the structural properties of GNNs.
The BK is provided as a fixed undirected graph, and each training sample consists of node features defined over this graph.
The task is graph classification: informed models receive both the BK graph and sample-specific node features as input, while uninformed models operate solely on the node features.

\subsubsection{Data Outline}
The overall structure of the synthetic setting aligns with that used in \cite{liMultimodalGraphNeural2024} and MPK-GNN \cite{xiaoGraphNeuralNetworks2023}.
For each sample (i.e., each row in the dataset), a graph is constructed by assigning the sample’s values as node features on a fixed underlying topology defined by the \ac{BK} graph.
Thus, all input graphs share the same structure, while the node features vary across samples.

\subsubsection{Graph Structure}
The graph is composed of equally sized, fully connected clusters, each representing a distinct class.
Nodes correspond to features in the training data, and edges indicate that two features are drawn from the same distribution.
There are no connections between clusters, ensuring that only intra-cluster structure carries useful information.

\subsubsection{Node Features}
To ensure the BK graph contributes meaningful information, node features are drawn from overlapping distributions, making classification based on features alone difficult.
Many nodes share identical distributions, adding redundancy, while only a sample-specific small subset carries discriminative signals.
The BK graph connects nodes from the same distribution, enabling GNNs to propagate useful information from informative nodes to less informative ones.
This structure helps separate classes, whereas uninformed models lack the means to identify distributional relationships.

\subsubsection{Definitions}
Formally, we define a synthetic dataset as:
\begin{equation}
    \label{eq:syn_dataset}
    \mathcal{D}(C,M,N) = (G, \mathcal{X}, y)
\end{equation}
where $G = (V, E)$ is a graph, $\mathcal{X} \in \mathbb{R}^{CN \times CM}$ represents the corresponding training data, and $y \in \{0, 1, \dots, C-1\}^{CN}$ denotes the labels.
The dataset is characterized by the parameters $C$, $M$ and $N$, which denote the number of clusters (or classes), nodes per cluster and samples per class, respectively.

\subsubsection{Graph Construction}
\label{sec:syn_graph_const}

Let $\mathcal{C} = \{0, 1, \dots, C-1\}$ be the set of cluster indices.
For each cluster $c \in \mathcal{C}$, the set of nodes in this cluster is defined as:
\begin{equation}
  V_c = \{cM, cM+1, \dots, cM+M-1\}
\end{equation}
Thus, each cluster contains $M$ nodes, which are identified by their index.
Similarly, the set of edges within the cluster is defined as:
\begin{equation}
  E_c = \{uv \mid u, v \in V_c, u \neq v\}
\end{equation}
Each cluster is fully connected, meaning there is an edge between every pair of nodes within a cluster, but no self-connections.

Finally, the graph $G = (V, E)$ is defined as:
\begin{equation}
  V = \bigcup_{c \in \mathcal{C}} V_c, \qquad E = \bigcup_{c \in \mathcal{C}} E_c
\end{equation}

The graph thus consists of $CM$ nodes, with no edges between clusters.
Each node has a degree of $M-1$, connecting to all other nodes within the same cluster.
Each cluster contains $M(M-1)/2$ edges, resulting in a total of $CM(M-1)/2$ edges in the graph.

\subsubsection{Training Data Generation}
\label{sec:syn_training_data_gen}

For data generation, we use two skew normal distributions ---that we will refer to as the active and inactive distributions--- which are symmetric to each other:
\begin{equation}
  X_\top \sim \mathcal{SN}\left(\frac{\Delta_\xi}{2}, \omega, \alpha\right), \qquad
  X_\bot \sim \mathcal{SN}\left(-\frac{\Delta_\xi}{2}, \omega, -\alpha\right)
\end{equation}
where $\Delta_\xi \in \mathbb{R}$ denotes the distance between locations, $\omega$ controls the spread and $\alpha$ controls the skewness.
These parameters determine the amount of overlap $\Omega$ between the distributions.

\begin{figure}[htbp]
    \centering
    \begin{subfigure}[T]{0.48\columnwidth}
        \includegraphics[width=\linewidth]{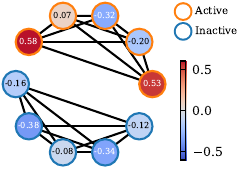}
        \caption{
            A sample with $C=2$, $M=5$ and the node features of the top cluster are from the active distribution.
        }
        \label{fig:syn_graph}
    \end{subfigure}
    \begin{subfigure}[T]{0.48\columnwidth}
        \includegraphics[width=\linewidth]{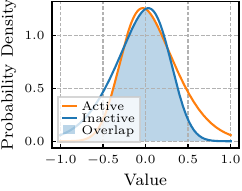}
        \caption{
            $\Delta_\xi=-0.57$, $\omega=0.5$, $\alpha=1.8$, $\Omega \approx0.904$
        }
        \label{fig:syn_dists}
    \end{subfigure}
    \caption{
        An example graph from the synthetic setting (a) and the corresponding active/inactive distributions (b).
    }
    \Description{
        An example graph and the active and inactive distributions.
    }
    \label{fig:syn_aux}
\end{figure}

For a graph $G = (V, E)$ assigned to class $c$, node features within the corresponding cluster are drawn from the active distribution, while features of all remaining nodes are drawn from the inactive distribution.
The classification task is to identify the cluster associated with the active distribution.
\Cref{fig:syn_aux} shows an example graph and the corresponding active/inactive distributions.

\section{Benchmark Application: Cancer Subtype Classification}
\label{sec:results_real}

In this section, we present the results obtained on real datasets for cancer subtype classification task.
This real-world task serves as a benchmark to assess 
\begin{inparaenum}[(i)]
    \item how much the BK improves classification performance compared to uninformed baselines, and
    \item how BK-aware methods respond to imperfections in the BK.
\end{inparaenum}

\paragraph{Methods}
As introduced in \cref{sec:bg_csc}, Li24 \cite{liMultimodalGraphNeural2024} and \ac{MPK-GNN} \cite{xiaoGraphNeuralNetworks2023} models are selected for this benchmark application.
Both models are trained and evaluated on the \ac{PANCAN} dataset, while Li24 model is also trained and evaluated on the \ac{BRCA} dataset.
In addition to these models, uninformed baselines, namely \ac{MLP}, \ac{SVM} and \ac{LogReg} are also tested.
These baselines only utilize the multi-omics data without the \ac{BK} graphs.
For this task, the original publications are closely followed using publicly available implementations, model architectures and hyperparameters.

The results are categorized into two major findings, namely 
\begin{inparaenum}
    \item uninformed baselines outperform BK-aware GNN methods (\cref{sec:res_real_uninf_are_superior}), and
    \item BK-graph structure is hardly used (\cref{sec:res_real_bk_utilization}).
\end{inparaenum}

\subsection{Uninformed Baselines Outperform Informed Models}
\label{sec:res_real_uninf_are_superior}

\paragraph{LogReg: An Overlooked Strong Baseline}
The average classification accuracies are shown in \cref{fig:res_real_rem,fig:res_real_add}.
A key finding across almost all settings is that uninformed models consistently outperform informed ones.
Specifically, the \ac{LogReg} model with L1 regularization surpasses both Li24 and \ac{MPK-GNN} on both datasets—a surprising result, given that the informed models are explicitly designed to leverage \ac{BK} for improved classification.
Although each informed model is evaluated against uninformed baselines in its original paper---Li24 against an \ac{MLP}, and \ac{MPK-GNN} against \ac{SVM}, \ac{RF}, and \ac{kNN}---\ac{LogReg} is not included among them, yet proves to be a particularly strong baseline in our experiments.

\paragraph{Architecture Discrepancy in Li24}
A potential methodological concern with the Li24 model is that its parallel \ac{MLP} uses a different architecture than the baseline \ac{MLP} it is compared to—two hidden layers versus three.
While the reason for this discrepancy is not explained in the original work, we test both configurations in our experiments.
The performance of the Li24 GAT model closely matches that of the two-hidden-layer \ac{MLP} with $82.2\% \pm 2.6\%$ on PANCAN (\cref{fig:li_pancan_rem}) and with $84.7\% \pm 2\%$ accuracy on BRCA (\cref{fig:li_brca_rem}).
Specifically for BRCA dataset, this suggests that the model’s performance gains likely stem from architectural differences in the \ac{MLP} rather than effective use of the BK graph.

Overall, these results challenge the assumption that incorporating \ac{BK} via graph-based methods reliably improves classification accuracy in this setting.

\subsection{Response to Imperfect BK: Limited Use of Graph Structure}
\label{sec:res_real_bk_utilization}

Another key observation across most perturbation types (\cref{fig:res_real_add,fig:res_real_rem,fig:li_brca_noise}) is that informed models exhibit either minimal sensitivity or erratic performance fluctuations.
These findings suggest that the tested \ac{GNN} models do not effectively utilize \ac{BK}.
We now examine how the models perform under different perturbations.

\subsubsection{Removing Random Edges}

\begin{figure*}[htbp]
    \centering
    \begin{subfigure}[T]{0.3\fulltextwidth}
        \includegraphics[width=\linewidth]{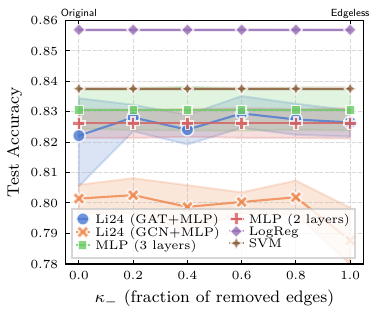}
        \caption{Li24 on PANCAN}
        \label{fig:li_pancan_rem}
    \end{subfigure}
    \begin{subfigure}[T]{0.3\fulltextwidth}
        \includegraphics[width=\linewidth]{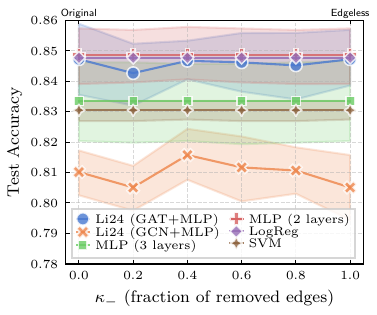}
        \caption{Li24 on BRCA}
        \label{fig:li_brca_rem}
    \end{subfigure}
    \begin{subfigure}[T]{0.3\fulltextwidth}
        \includegraphics[width=\linewidth]{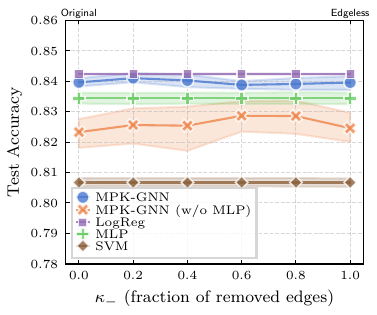}
        \caption{MPK-GNN on PANCAN}
        \label{fig:mpkgnn_pancan_rem}
    \end{subfigure}
    \caption{
        Average accuracy (over 10 runs) of Li24 and \ac{MPK-GNN} models compared to uninformed baselines on PANCAN and BRCA datasets for varying $\kappa_{-}$ values.
        Error bands indicate $95\%$ confidence interval.
    }
    \Description{
        Average accuracy curves across different values of perturbation severity values, in this case the fraction of removed edges.
        In all cases, the uninformed methods overperform the informed ones, and the informed methods show minimal response to the edge removals.
    }
    \label{fig:res_real_rem}
\end{figure*}

The average accuracy results for random edge removal perturbation are shown in \cref{fig:res_real_rem}.

\paragraph{Minimal Sensitivity to Edge Removals}
All models show minimal sensitivity to edge removals.
Remarkably, even when all edges are removed, model performance remains largely unchanged.
This suggests that the graph topology intended as \ac{BK} does not meaningfully influence the learning process of these \ac{GNN} models.
This trend holds across nearly all datasets and model variants.
The sole exception is the Li24 \ac{GCN} model on the \ac{PANCAN} dataset, which shows a modest performance drop of $1.4\%$ when all edges are removed.

\subsubsection{Adding Random Edges}

\begin{figure*}[htbp]
    \centering
    \begin{subfigure}[T]{0.3\fulltextwidth}
        \includegraphics[width=\linewidth]{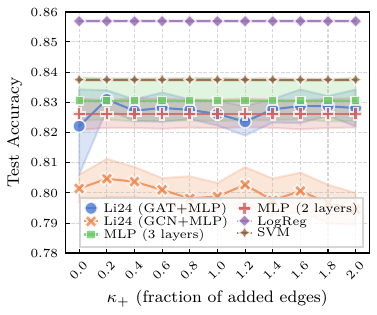}
        \caption{Li24 on PANCAN}
        \label{fig:li_pancan_add}
    \end{subfigure}
    \begin{subfigure}[T]{0.3\fulltextwidth}
        \includegraphics[width=\linewidth]{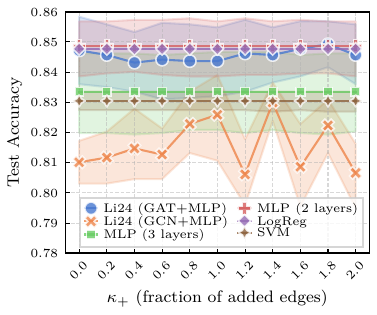}
        \caption{Li24 on BRCA}
        \label{fig:li_brca_add}
    \end{subfigure}
    \begin{subfigure}[T]{0.3\fulltextwidth}
        \includegraphics[width=\linewidth]{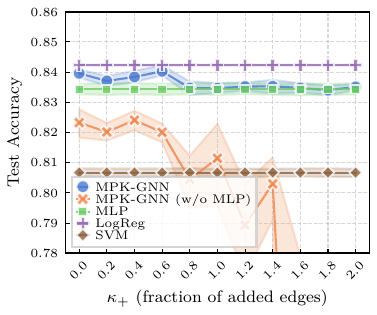}
        \caption{MPK-GNN on PANCAN}
        \label{fig:mpkgnn_pancan_add}
    \end{subfigure}
    \caption{
        Average accuracy (over 10 runs) of Li24 and MPK-GNN models compared to uninformed baselines on PANCAN and BRCA datasets for varying $\kappa_{+}$ values.
        Error bands indicate $95\%$ confidence interval.
    }
    \Description{
        Average accuracy curves across different values of perturbation severity values, in this case the fraction of added edges.
        In all cases, the uninformed methods overperform the informed ones.
    }
    \label{fig:res_real_add}
\end{figure*}

The average accuracy results for random edge addition perturbation are shown in \cref{fig:res_real_add}.
Overall, models show a slightly more pronounced response to edge addition.

\paragraph{Li24}
For Li24 on the PANCAN dataset (\cref{fig:li_pancan_add}), both GCN and GAT variants remain stable, with performance nearly constant across $\kappa_{+}$ levels.
On the BRCA dataset (\cref{fig:li_brca_add}), performance of the GAT variant initially declines slightly but then recovers, eventually surpassing the \ac{LogReg} baseline by $0.1\%$ accuracy at $\kappa_{+}=1.8$.
The GCN variant on the other hand, shows fluctuations after $\kappa_{+}=0.6$, barely reaching \ac{SVM} at $\kappa_{+}=1.4$.

\paragraph{MPK-GNN}
For MPK-GNN, we observe a sharp decline in performance of the variant without the MLP component at $\kappa_{+}=1.6$ (\cref{fig:mpkgnn_pancan_add}).
Beyond this point, most runs fail to learn meaningful representations, evidenced by stagnant loss curves and performance approaching random guessing levels.
We attribute this to over-smoothing, where node embeddings become overly similar \cite{chenMeasuringRelievingOversmoothing2019, liuDeeperGraphNeural2020}.
Since adding edges effectively increases the receptive field, similar to stacking more GNN layers, nodes aggregate not only relevant but also noisy signals from distant parts of the graph.
As the embeddings converge, the model loses discriminative power, which results in degraded accuracy.
In contrast, when the MLP is included, it appears to mitigate the GNN's collapse.

\subsubsection{Noise on Edge Weights}

\begin{figure}[htbp]
    \centering
    \begin{subfigure}[T]{\columnwidth}
        \includegraphics[width=\linewidth]{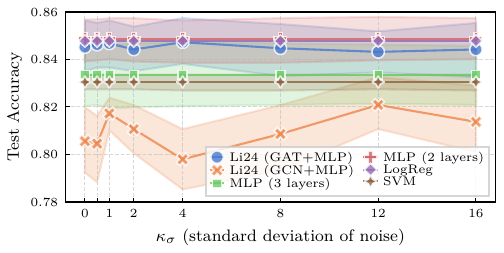}
        \caption{Variant: Remove Negative Edges}
        \label{fig:li_brca_noise_nonew}
    \end{subfigure}
    \\
    \begin{subfigure}[T]{\columnwidth}
        \includegraphics[width=\linewidth]{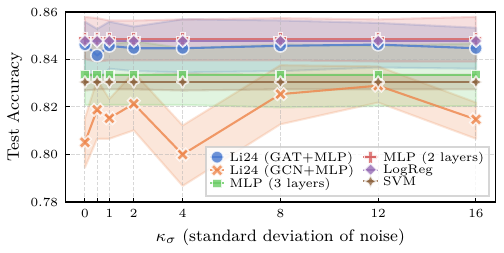}
        \caption{Variant: Replace Negative Edges}
        \label{fig:li_brca_noise_new}
    \end{subfigure}
    \caption{
        Average accuracy (10 runs) of Li24 compared to uninformed baselines on \ac{BRCA} dataset for varying $\kappa_{\sigma}$ values.
    }
    \Description{
        Average accuracy curves across different values of perturbation severity values, in this case the noise on edge weights.
        In all cases, the uninformed methods overperform the informed ones.
    }
    \label{fig:li_brca_noise}
\end{figure}

This perturbation type, described in \cref{sec:pert_noise}, is applied to Li24 model on the \ac{BRCA} dataset.
The average accuracy results are given in \cref{fig:li_brca_noise}.

\paragraph{Minimal Sensitivity to Edge Weight Noise}
Li24 with GAT again shows only minor sensitivity to perturbation, whereas the GCN variant displays more variability across perturbation levels.
Interestingly, GCN models tend to exhibit slight performance improvements as noise increases, regardless of whether negative-weight edges are clipped (\cref{fig:li_brca_noise_nonew}) or replaced with new random edges (\cref{fig:li_brca_noise_new}).
One possible explanation is that pruning predominantly removes edges with low evidence of interaction, thereby reducing the amount of irrelevant information aggregated from loosely related nodes.
Alternatively, the addition of new random edges expands the receptive field, enabling nodes to incorporate information from previously disconnected parts of the graph.

\section{Synthetic Setting: Informed Models are Better on Optimal Data}
\label{sec:results_synthetic}

We now investigate whether the surprising real-world results also hold when the BK is clearly informative.
First, we compare informed models to uninformed baselines on unperturbed data in \cref{sec:res_syn_unperturbed}, then examine their responses to various perturbations in \cref{sec:res_syn_perturbed}.
Dataset parameters are $C=2$, $M=16$, $N=600$ with $\Delta_\xi=-0.57$, $\omega=0.5$ and $\alpha=1.8$, which corresponds to a distributional overlap of $\Omega \approx 0.904$ as shown in \cref{fig:syn_dists}.

\subsection{No Perturbation}
\label{sec:res_syn_unperturbed}

We begin by testing informed and uninformed models on unperturbed data to assess whether the performance gap from the real-world setting persists when BK is clearly informative.
To complement this, we also evaluate a simple form of BK integration by running \ac{LogReg} and \ac{SVM} on the average node features of clusters.
This provides the classifiers with aggregated structural information without directly using the BK graph.
In addition to that, we run GNNs with GCN and GATv2 layers, Li24 model (GAT), Li24 with only the parallel MLP, and the uninformed baselines.

\begin{figure}[htbp]
    \centering
    \includegraphics[width=\columnwidth]{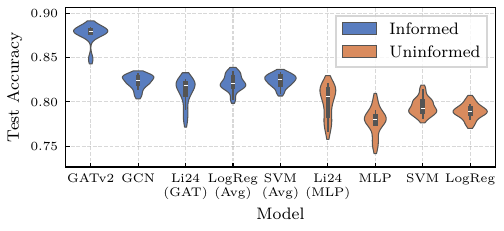}
    \caption{
        Informed and uninformed models tested on the synthetic data without perturbations.
    }
    \Description{
        Violin plots showing test accuracy distributions of informed and uninformed methods on the synthetic data.
        Overall, informed models perform better than uninformed ones.
    }
    \label{fig:syn_inf_uninf}
\end{figure}

Results in \cref{fig:syn_inf_uninf} show that informed models consistently outperform uninformed ones in the synthetic setting with no perturbations.
Even relatively simple informed models, such as \ac{SVM} and \ac{LogReg} applied to cluster-averaged features, achieve higher test accuracy than all uninformed baselines.
This demonstrates that \ac{BK} can offer meaningful predictive value, even with basic models.

\subsection{Response to Perturbations}
\label{sec:res_syn_perturbed}

\begin{figure*}[htbp]
    \centering
    \begin{subfigure}[T]{0.23\fulltextwidth}
        \includegraphics[width=\linewidth]{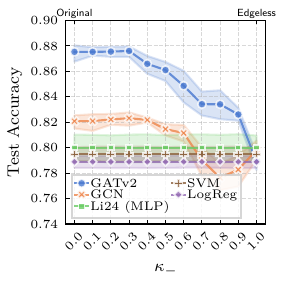}
        \caption{Removing Edges}
        \label{fig:syn_rem}
    \end{subfigure}
    \begin{subfigure}[T]{0.23\fulltextwidth}
        \includegraphics[width=\linewidth]{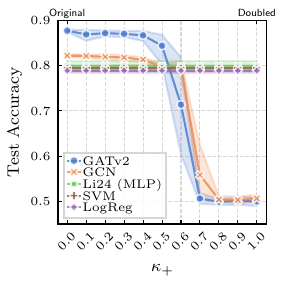}
        \caption{Adding Edges}
        \label{fig:syn_add}
    \end{subfigure}
    \begin{subfigure}[T]{0.23\fulltextwidth}
        \includegraphics[width=\linewidth]{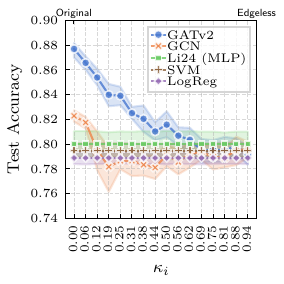}
        \caption{Isolating Nodes}
        \label{fig:syn_iso}
    \end{subfigure}
    \begin{subfigure}[T]{0.23\fulltextwidth}
        \includegraphics[width=\linewidth]{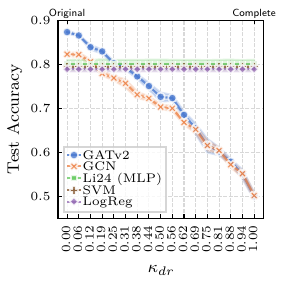}
        \caption{Detach-and-Rewire}
        \label{fig:syn_dr}
    \end{subfigure}
    \caption{
        Average accuracies across 10 runs on the synthetic dataset with different perturbations. 
    }
    \Description{
        Average accuracy curves on the synthetic data across different values of perturbation severity values.
        Informed methods show more apparent response to the perturbations on the synthetic data.
    }
    \label{fig:syn_pert}
\end{figure*}

The average accuracy curves for varying $\kappa$ values are shown in \cref{fig:syn_pert}.
Overall, informed GNNs show robustness to graph perturbations, though different modifications trigger distinct responses and affect the extent of robustness, which is discussed in \cref{sec:res_syn_rem,sec:res_syn_add,sec:res_syn_iso,sec:res_syn_dr}.

\subsubsection{Removing Random Edges}
\label{sec:res_syn_rem}

An immediate observation from \cref{fig:syn_rem} is that both GCN and GATv2 exhibit robustness to edge removal, with performance remaining stable even when up to $30\%$ of edges are removed.
We observe a decline after this point, however even with $90\%$ of edges randomly removed, GATv2 maintains a $2.6\%$ performance gain over the baseline MLP.
The accuracy of GCN falls below the baseline when $\kappa_{-}=0.7$, and for a fully disconnected graph, both models converge to baseline-level performance.

\begin{figure}[htbp]
    \centering
    \begin{subfigure}[T]{0.48\columnwidth}
        \includegraphics[width=\linewidth]{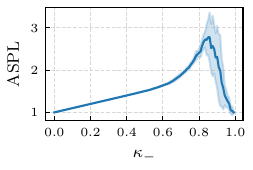}
        \caption{
            \Ac{ASPL} of a cluster in the BK graph for varying values of $\kappa_{-}$.
        }
        \label{fig:aspl_rem}
    \end{subfigure}
    \begin{subfigure}[T]{0.48\columnwidth}
        \includegraphics[width=\linewidth]{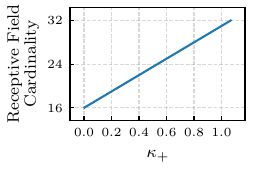}
        \caption{
            Average 1-hop receptive field cardinality of nodes in a cluster.
        }
        \label{fig:rf_add}
    \end{subfigure}
    \caption{
        \Ac{ASPL} and average receptive field cardinality values in a cluster for random edge removal and addition perturbations, respectively.
    }
    \Description{
        Average shortest path length and receptive field cardinalities in a cluster.
        ASPL increases slowly until $80\%$ removal, and the cardinality is linearly increasing from 16 to 32 with more added edges.
    }
    \label{fig:add_rem_aux}
\end{figure}

We hypothesize that this robustness stems from the graph’s local structure: even after edge removals, the \ac{ASPL} within clusters remains low (\cref{fig:aspl_rem}).
Originally, clusters are fully connected, with path lengths of 1 between nodes.
After deletion, most nodes remain reachable within three hops—the depth of the GNNs—enabling effective propagation of key node embeddings.
An exception is the disconnected regions caused by removals; while they still show short \ac{ASPL}, they may miss important information due to lost connectivity.

In the case of full edge removal, only self-loops remain.
While self-loops are not part of the definition in \cref{sec:syn_graph_const}, they are important for most GNN layers like GCN and are typically added by default in common frameworks.
Here, each node updates its embedding based solely on its own previous embedding.
For GATv2, this results in an attention weight of 1 for the self-loop due to the softmax over a single edge.
The GNN thus degenerates into a local transformation.
Subsequent linear layers applied to these node embeddings closely resemble those in the MLP, explaining the convergence in performance.

\subsubsection{Adding Random Edges}
\label{sec:res_syn_add}

Despite all added edges connecting nodes from different clusters, both informed models remain stable even when up to $40\%$ additional random edges are introduced (\cref{fig:syn_add}).
However, performance drops sharply beyond $\kappa_{+}\geq 0.5$.
At this point, some training runs still converge to meaningful representations, while others fail entirely, with accuracy plateauing near chance level.
For $\kappa_{+}\geq 0.8$, all runs yield accuracy around $50\%$.
The training loss fails to decrease in these cases, indicating that the models are unable to learn useful parameters.
Interestingly, GCN shows slightly greater resilience than GATv2 under these conditions.
This degradation in performance with increasing edge noise mirrors trends observed in MPK-GNN (Figure~\ref{fig:mpkgnn_pancan_add}).
We attribute this decline in performance to over-smoothing: as more inter-cluster edges are added, nodes increasingly aggregate information from the opposite cluster.
This expansion of the receptive field, shown in \cref{fig:rf_add}, causes node embeddings across clusters to become overly similar, thereby impairing classification performance.

\subsubsection{Isolating Nodes}
\label{sec:res_syn_iso}

Isolating nodes is similar to edge removal, but here edges are removed in a more systematic manner.
Notably, isolating just two nodes causes GCN’s accuracy to fall below the baseline, while GATv2 retains a performance advantage until $\kappa_{i}=11/16$ (\cref{fig:syn_iso}).
In contrast with random edge removal, isolating nodes leads to a steeper decline in accuracy for both informed models, converging to the baseline level when $\kappa_{i} > 11/16$.

Isolating a node removes all its edges at once, preventing it from contributing to message passing. 
In contrast, random edge removal often preserves some intra-cluster connectivity, allowing partial information flow.
Isolated nodes are only integrated with others during the final concatenation and classification stages, which explains the sharper performance degradation.

\subsubsection{Detach-and-Rewire}
\label{sec:res_syn_dr}

Among all perturbation strategies, detach-and-rewire has the most pronounced impact on model performance.
GCN drops below the baseline after only three nodes are rewired ($\kappa_{dr}=3/16$), while GATv2 maintains an advantage until $\kappa_{dr}=5/16$.
At full perturbation ($\kappa_{dr}=1$), the graph resembles a fully connected structure—similar to that produced by edge addition, resulting in accuracy dropping to around $50\%$.
However, in contrast to edge addition, the decline in performance is more gradual, with accuracy decreasing smoothly as more nodes are rewired.

The key distinction lies in how the graph structure evolves.
While adding even a few edges quickly connects the two clusters, rewiring nodes preserves the existence of two distinct groups.
As $\kappa_{dr}$ increases, nodes from the second (smaller) cluster are moved into the first (larger) one.
Full connectivity is only achieved at the final perturbation stage.
Until then, noise from the first cluster does not reach the second cluster.
However, nodes that are moved into the first cluster introduce noise there, since their features originate from the second cluster’s distribution.
This gradually disrupts the representation quality within the first cluster.
We hypothesize that over-smoothing primarily affects the first cluster in this scenario, while the second remains relatively unaffected.
As long as the second cluster retains some informative node embeddings, this may explain the smoother performance degradation compared to the more abrupt collapse observed in edge addition.

\section{Discussion}
\label{sec:discussion}

While integrating BK into GNNs is often expected to boost performance, our findings reveal a more nuanced reality.
In the real-world cancer subtype classification task, two surprising findings emerge:
\begin{inparaenum}[(i)]
    \item uninformed baselines outperform BK-informed GNNs, and
    \item BK-informed GNNs appear to make minimal use of the BK graph.
\end{inparaenum}
An immediate verification regarding these findings is testing these models on a controlled setting, where the BK is clearly informative.
Consequently, in the synthetic setting, we observed 
\begin{inparaenum}[(i)]
    \item informed models perform better than the uninformed baselines, and
    \item a decline in performance as the perturbation severity increases, which shows that the BK is actually utilized.
\end{inparaenum}

Given these, we now discuss our findings across three themes: benefit, robustness, and implications.

\subsection{When Do BK Graphs Improve GNN Model Performance?}

The findings of this study on the cancer subtype classification task provide a counter-example to the common assumption that incorporating BK in GNNs reliably improves performance.
In most cases, uninformed baselines outperform their informed counterparts, and GNNs show minimal sensitivity to BK perturbations.
This is especially notable for STRINGDB, which meets prior criteria for useful BK \cite{bertinAnalysisGeneInteraction2020} but still fails to improve MPK-GNN performance.

In contrast, the synthetic setting demonstrates that GNNs can effectively leverage BK when conditions are optimal.
Models informed by BK consistently outperform uninformed baselines, and this advantage holds across a range of dataset perturbations.
Unlike in the real-world experiments, GNNs in the synthetic setting exhibit clear and systematic responses to perturbation, indicating that they meaningfully incorporate BK into the learning process.

Another notable observation is that attention mechanisms can enhance the utility of BK by selectively amplifying the propagation of relevant node features.
This effect is evident in the synthetic experiments, where the GATv2 model outperforms GCN when leveraging BK (\cref{fig:syn_inf_uninf}).
However, attention alone does not guarantee improved performance.
For instance, the Li24 GAT model, which also employs attention, fails to surpass baseline models (\cref{fig:li_pancan_rem,fig:li_brca_rem}), indicating that additional factors influence how effectively BK is integrated.

A key question remains: what properties must BK graphs, model architectures, and datasets have for GNNs to effectively leverage BK?
The synthetic setting illustrates when BK is helpful but does not reveal which conditions are necessary or sufficient.
These graphs feature strong clustering and skewed feature distributions, with dense intra-cluster links and no inter-cluster edges—conditions that help informative signals propagate cleanly.
Yet, it is unclear which of these traits matter most.
Closing this gap is essential for harnessing BK in real-world tasks. While this study identifies conditions where BK is beneficial, further work is needed to understand why real-world methods often fail to benefit from it.

\subsection{How Robust Are GNNs to Errors in BK?}

Beyond measuring performance gains, this study also assessed GNN robustness to errors in BK.
In real-world settings, the tested models appeared highly robust to BK perturbations, yet this likely reflects limited use of the BK structure rather than true resilience.
In contrast, the synthetic setting offers clearer insight, as BK meaningfully impacts performance and allows detailed analysis of how different error types affect robustness.

The results show that GNNs are generally more resilient to missing than incorrect knowledge, especially in extreme cases as seen in \cref{fig:syn_pert}.
Removing most edges or isolating nodes reduces the benefits of BK but does not drastically harm performance, as the model behavior approaches that of an MLP on sparse graphs.
In contrast, introducing substantial incorrect knowledge through added edges or node rewiring can significantly degrade performance, likely due to over-smoothing, where message passing across incorrect connections blurs critical feature distinctions.

Moreover, while GNNs can tolerate moderate levels of random noise (e.g., randomly deleting up to $30\%$ or adding up to $40\%$ of edges in the synthetic setting), they are far more sensitive to systematic errors.
Even minor targeted perturbations such as selectively isolating or rewiring nodes lead to steady performance declines, with losses accumulating as the perturbation gets more severe.

These findings highlight the importance of understanding and validating the specific errors in real-world BK graphs such as the \ac{GGI}/\ac{PPI} networks.
While some noise has minimal impact, systematic biases can severely impair model performance, emphasizing the need for rigorous evaluation and possible correction before integrating BK into GNNs.

\subsection{Implications}

These findings carry important implications for developing GNN models that utilize BK.
First, performance gains over baseline models should not be assumed to reflect meaningful use of BK.
Perturbation analyses introducing different types of errors into the BK graph are essential to verify whether the model genuinely relies on the knowledge.
Second, systematic errors in BK can have strong negative effects.
Researchers should evaluate potential biases in their BK sources and consider mitigation strategies, such as weighting or filtering low-confidence edges.
Finally, sparse, high-quality BK is often more effective than dense, noisy graphs.
Dense graphs are more susceptible to the harms of erroneous connections.
While BK can enhance GNN performance, its benefits require careful validation and thoughtful integration.

\section{Conclusion and Future Work}
\label{sec:conclusion}

This study takes a first step toward understanding when BK graphs benefit GNNs and how robust these models are to imperfections.
In a real-world cancer classification task, state-of-the-art GNNs using BK show no clear advantage over uninformed baselines and are largely insensitive to substantial perturbations.
To interpret these findings, we introduced a controlled synthetic setting with informative BK, where GNNs succeed, provided their architecture aligns well with the BK structure.

These findings suggest that current approaches may not fully exploit the potential of BK, and future research should focus on identifying conditions under which BK is most beneficial for GNNs.
Graph-theoretic measures may help characterize properties of biological networks that make them beneficial in addition to the data available.
Further directions include designing adaptive GNNs, creating metrics to assess BK informativeness, and extending this evaluation framework with more targeted perturbation techniques.

\begin{acks}
    Funding for this study was provided by the BMBF (01IS22077).
\end{acks}

\balance
\bibliographystyle{abbrvnat}
\bibliography{main.bib}

\end{document}

%% file: glossary.tex
\newacronym{ASPL}{ASPL}{average shortest path length}
\newacronym{BK}{BK}{background knowledge}
\newacronym{BRCA}{BRCA}{Breast Invasive Carcinoma}
\newacronym{GAT}{GAT}{graph attention network}
\newacronym{GCN}{GCN}{graph convolutional network}
\newacronym{GGI}{GGI}{gene-gene interaction}
\newacronym{GNN}{GNN}{graph neural network}
\newacronym{iEN}{iEN}{immunological Elastic-Net}
\newacronym{IML}{IML}{Informed Machine Learning}
\newacronym{kNN}{kNN}{k-nearest neighbors}
\newacronym{LogReg}{LogReg}{logistic regression}
\newacronym{ML}{ML}{Machine Learning}
\newacronym{MLP}{MLP}{multi-layer perceptron}
\newacronym{MPK-GNN}{MPK-GNN}{Multiple Prior Knowledge GNN}
\newacronym{ODE}{ODE}{ordinary differential equation}
\newacronym{PANCAN}{PANCAN}{Pan-Cancer}
\newacronym{PPI}{PPI}{protein-protein interaction}
\newacronym{RF}{RF}{random forest}
\newacronym{SVM}{SVM}{support vector machine}
\newacronym{TCGA}{TCGA}{The Cancer Genome Atlas}